\documentclass{article}

\usepackage{svg}

\usepackage[final]{neurips_2024}

\usepackage[utf8]{inputenc} 
\usepackage[T1]{fontenc}    
\usepackage{hyperref}       
\usepackage{url}            
\usepackage{booktabs}       
\usepackage{amsfonts}       
\usepackage{nicefrac}       
\usepackage{microtype}      
\usepackage{xcolor}         
\usepackage{graphicx}
\usepackage{arydshln}
\usepackage{subfigure}
\usepackage{subcaption}

\title{Chitrarth: Bridging Vision and Language \\ for a Billion People}

\author{Shaharukh Khan, Ayush Tarun, Abhinav Ravi *, Ali Faraz, Akshat Patidar \\ \textbf{Praveen Pokala *, Anagha Bhangare, Raja Kolla, Chandra Khatri *, Shubham Agarwal *} \\ \\
Krutrim AI, Bangalore, India\\
\texttt{* Senior Contributors}\\
\textsuperscript{Contact: \{shaharukh.khan, abhinav.ravi, shubham.agarwal1\}@olakrutrim.com} 
}

\begin{document}

\maketitle

\begin{abstract}
Recent multimodal foundation models are primarily trained on English or high resource European language data, which hinders their applicability to other medium and low-resource languages.
To address this limitation, we introduce \textit{Chitrarth} (Chitra: Image; Artha: Meaning), an inclusive Vision-Language Model (VLM), specifically targeting the rich linguistic diversity and visual reasoning across 10 prominent Indian languages. Our model effectively integrates a state-of-the-art (SOTA) multilingual Large Language Model (LLM) with a vision module, primarily trained on multilingual image-text data.
Furthermore, we also introduce BharatBench, a comprehensive framework for evaluating
VLMs across various Indian languages, ultimately contributing to more diverse and effective AI systems. Our model achieves SOTA results for benchmarks across low resource languages while retaining its efficiency in English. Through our research, we aim to set new benchmarks in multilingual-multimodal capabilities, offering substantial improvements over existing models and establishing a foundation to facilitate future advancements in this arena. 
\end{abstract}

\section{Introduction}




With the success and demonstrated effectiveness  of \textit{Visual instruction tuning} ~\citep{liu2024visual,liu2024improved}, recent years witnessed a surge of
interest in developing general purpose multimodal conversational agents. These unified foundation models excel at algorithmic reasoning and generic perception tasks like image captioning, visual question answering, text-based image retrieval, etc.~\citep{lu2024deepseek,laurenccon2024matters,tong2024cambrian,xue2024xgen}, and more specialized frameworks, for instance, converting Scalable Vector Graphics (SVGs) to code~\citep{rodriguez2023starvector}. Often, these models rely on pre-trained Large Language Models (LLMs)~\citep{brown2020language,touvron2023llama,achiam2023gpt,chiang2023vicuna,touvron2023llama2,team2023gemini,jiang2024mixtral, team2024gemma, dubey2024llama} as the transformer~\citep{vaswani2017attention} backbones, primarily trained on English or high resource European languages.  

This work is driven by two main motivations: 1. \textit{Language diversity gap:} Most Vision Language Models (VLMs) are predominantly trained on English datasets, overlooking the linguistic needs of non-English languages, particularly from the Indian subcontinent. 2. \textit{Lack of low resource language benchmarks:} Absence of corresponding VLM benchmarks hinders the progress for these low resource Indic languages. We aim to address these issues through our research and serve a broader audience, encompassing billions of people.
 

Few LLMs have been developed specifically for Indic languages ~\citep{gala2024airavata, NavarasaTeluguLLMLabs, balachandran2023tamilllama, kohli2023building}, most of which extend and fine-tune text-only English-centric LLMs. Naturally, they fail to fully capture the nuances of the language, with the exception of models like~\citep{team2024krutrim,bendale2024sutra}, trained from scratch. Our model builds upon the recent success of Krutrim LLM~\citep{team2024krutrim}, which supports English and 10 other languages including Hindi, Bengali, Telugu, Tamil, Marathi, Gujarati, Kannada, Malayalam, Odia, and Assamese, representing a significant portion of the cultural and linguistic diversity in India. 

Another key challenge is the limited availability of low resource data with Indic languages significantly under-represented in Common Crawl despite India (or Bharat) making up 18\% of the global population. For instance, Hindi, in spite of being the third most spoken, does not appear among the top 20 languages~\citep{buck2014n,penedo2023refinedweb}. To enhance our model's cross-lingual generalization abilities, we translate the open-source multimodal training datasets into the 10 Indic languages natively supported by the backbone LLM. Developing this multilingual dataset is a substantial endeavor aimed at addressing the disparity between high-resource 
and relatively low-resource Indian languages in the context of vision-language models.

\begin{figure}[htbp]
  \centering
\centerline{\includegraphics[scale=0.27]{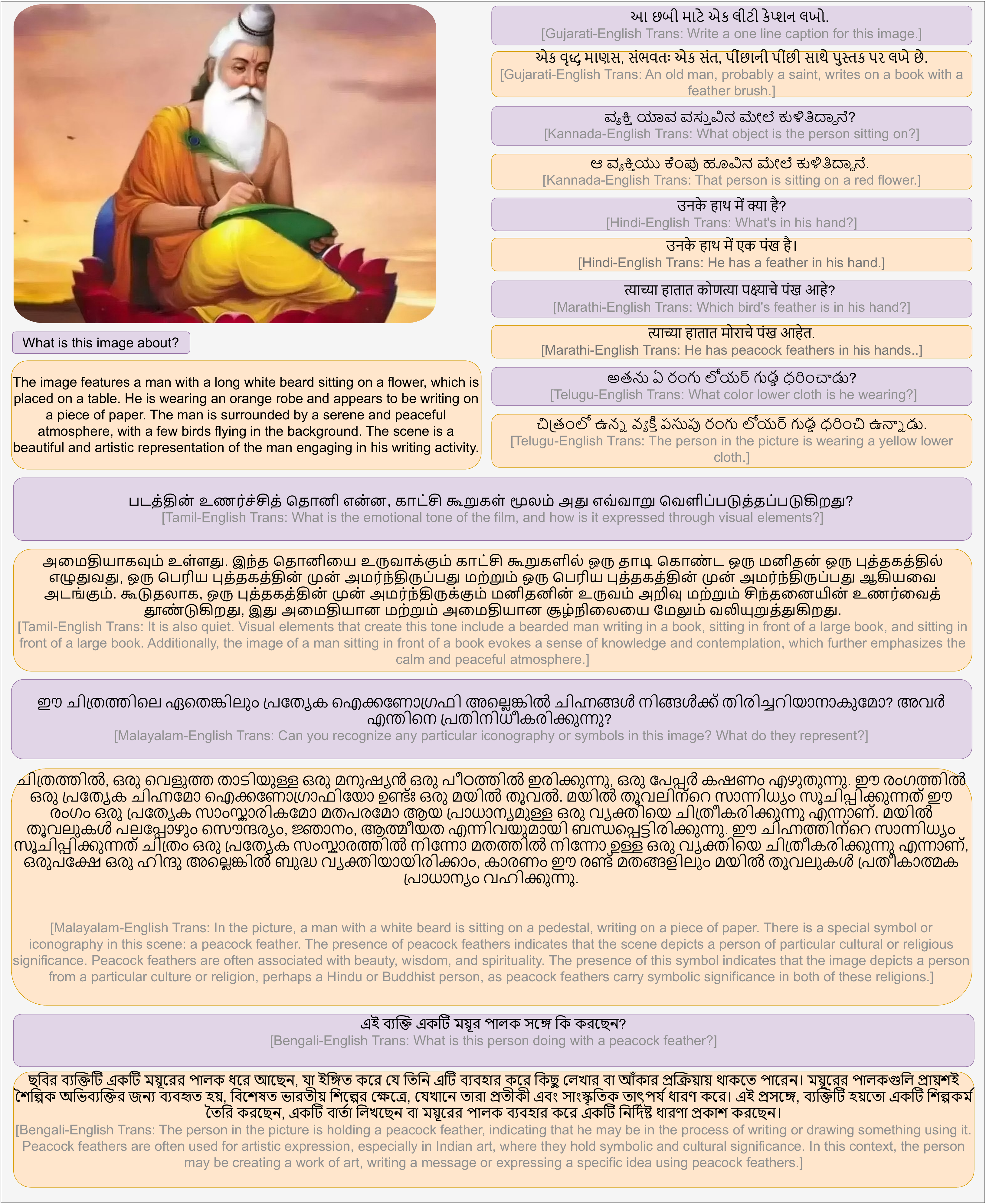}}
\caption{\textbf{Multi lingual capability of Chitrarth model across major Indian languages.} For the same underlying image, we present question-answer pairs in English and several Indian languages - Gujarati, Kannada, Hindi, Marathi, Telugu, Tamil, Malayalam, and Bengali (in order). Questions are highlighted in purple, and responses are shown in orange (provided with English translations). The model accurately understands and identifies the `image of a saint writing a book with a feather' and correctly addresses related questions in different languages.}
\label{fig:teaser}
\end{figure}

In this paper, we present our multimodal LLM, which employs the Krutrim multilingual LLM backbone ~\citep{team2024krutrim} in conjunction with a pre-trained visual image encoder ~\citep{alexey2020image}. Figure\ref{fig:teaser} demonstrates the multi-lingual capability of our model across major Indian languages. A brief summary of our contribution is provided below: 

\begin{itemize}

    \item We introduce Chitrarth (Chitra: Image; Artha: Meaning), a Multimodal LLM model which leverages images and language modalities for a range of visual tasks such as image captioning, visual question answering in the multilinugal context. We further present optimal training recipe including data composition and architecture configuration. 
    \item We also present \textit{BharatBench}, a comprehensive evaluation benchmark suite designed for 10 under-resourced Indic languages across 3 tasks, which we will make available to the research community upon acceptance.
    \item Finally, we evaluate Chitrarth and prior baselines on both existing English academic datasets as well as the proposed evaluation framework and demonstrate the effectiveness of our model, using different training strategies and ablations, achieving SOTA results on 3 out of 5 English datasets and propose benchmark results on the derived multi-lingual datasets.
\end{itemize}

The remainder of the paper is structured as: Section \ref{sec:relwork} reviews recent related research on VLMs. Section \ref{sec:model} provides a detailed description of our Chitrarth model with information about training data mix in Section \ref{sec:dataset}. Section \ref{sec:bench} introduces the BharatBench evaluation framework that we propose, while Section \ref{sec:exp} presents the experimental results. Finally, Section \ref{sec:conclusion} offers concluding remarks.

\section{Related Work}
\label{sec:relwork}

\subsection{English-centric VLMs}
Recent studies \citep{laurenccon2024matters, laurenccon2024building,tong2024cambrian} have investigated design strategies for multi-stage training pipelines in contemporary VLMs. Typically, these models rely on pre-trained LLMs; however, there are some exceptions where models are trained from scratch \citep{team2024chameleon, lu2024unified}. Prior works like Flamingo\citep{alayrac2022flamingo} leverage a Perceiver Resampler \citep{jaegle2021perceiver} to inject visual features into the language model through cross-attention, promoting quick adaptation to various tasks with few labeled examples. The LLaVA family models \citep{liu2024visual,liu2024improved}, including LLaVA-1.5 and LLaVA-1.6, demonstrated intriguing multimodal comprehension capabilities by integrating advanced language models with vision encoders through visual instruction tuning. 
PaliGemma~\citep{beyer2024paligemma}, optimized for tasks that require deep integration of visual and textual data, is designed to excel in scenarios where English is the primary language. Florence-2~\citep{xiao2024florence} focuses on handling diverse tasks from simple text prompts addressing the complexity of spatial hierarchy and semantic granularity. The Idefics family~\citep{laurenccon2024matters,laurenccon2024building} is focused on substantially enhancing capabilities around OCR, document interpretation and visual reasoning functionalities. CogVLM~\citep{wang2023cogvlm} drives an intricate fusion of language and vision features unlike other 
VLMs, which rely on the shallow alignment method. PALI models~\citep{chen2022pali} on the other hand explored contrastive pretraining and higher resolution training for the VLM tasks.

\subsection{Multi-lingual VLMs}
Qwen-VL \citep{bai2023qwen} is a multilingual VLM, trained on English and Chinese data, supporting diverse instructions and multi-image context analysis. InternVL 1.5 \citep{chen2024far} proposed an enhanced vision encoder and a superior bilingual dataset, i.e., English and Chinese. Phi-3 family ~\citep{abdin2024phi} offer multilingual, multimodal, and long-context support in 11 languages, including English, across the world but do not cover Indian languages. PALO \citep{maaz2024palo} is the closest VLM to our research, however supporting only 3 Indian languages Hindi, Urdu, and Bengali apart from the other high-to-medium resource language offerings. To our knowledge, no other open-source multimodal LLMs include low-resource Indic languages in the training mix. In contrast, our work introduces a multilingual VLM system that supports ten Indian languages.


\begin{figure}[htb]
\begin{minipage}[b]{1.0\linewidth}
  \centering
\centerline{\includegraphics[width=0.87\linewidth]{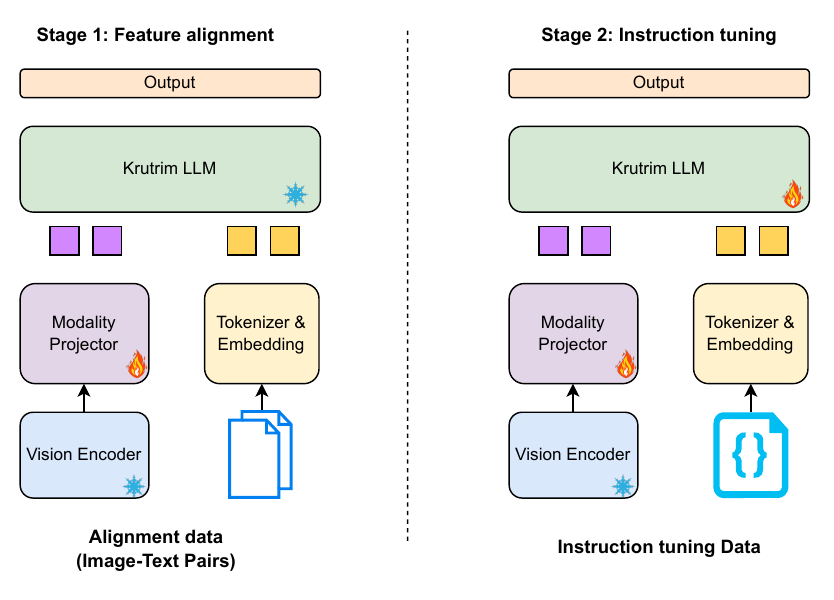}}
\end{minipage}
\caption{\textbf{Chitrarth model features a fully autoregressive architecture with a two-stage training process.} In Stage 1, the model is trained using images and their descriptions, aligning visual and linguistic embeddings through image-caption pairs. In Stage 2, model is fine-tuned on multimodal instruction-following and domain-specific academic datasets.}
\label{fig:model}
\end{figure}

\section{Chitrarth: A Multilingual Vision-Language Model}
\label{sec:model}

In this section, we outline the architecture of our proposed Chitrarth model. Chitrarth is an autoregressive VLM where the input image is tokenized into visual tokens, combined with textual instruction tokens and fed into the large language model (LLM). Inspired by the versatile and widely followed LLaVA~\citep{liu2024visual,liu2024improved} framework, our model incorporates several key components, as illustrated in Figure~\ref{fig:model}, where we use pre-trained Krutrim LLM~\citep{team2024krutrim} instead, as the autoregressive multi-lingual LLM backbone. 

For multimodal training, we start by encoding images using a vision encoder. The modality projection layer (adapter/connector) maps the vision embeddings into the LLM embedding space, producing a sequence of visual tokens. The multilingual LLM then generate responses based on these visual tokens. 
The Krutrim LLM~\citep{team2024krutrim} supports a context length of 4096 tokens, of which 576 tokens (14X14 patch size results in 729 tokens) are allocated for image representation after the modality projection. We explore different configurations for the projection layer, including a single-layer projection~\citep{liu2024visual,liu2024improved} and a two-layer MLP vision-language connector with non-linearity~\citep{liu2024visual}. Additionally, we experiment with various vision encoders, including the pre-trained CLIP ViT-L/14@336px~\citep{radford2021learning} and SigLIP-SO400M~\citep{zhai2023sigmoidlosslanguageimage}. Our model is trained in multiple stages:

\textbf{Stage 1: Pre-Training (PT) for Feature Alignment.} In this stage, we conduct pre-training using image-text pairs, with the projector layer being trained while keeping the vision encoder and LLM fixed. Each sample is treated as a single-turn conversational instruction for tuning.

\textbf{Stage 2: Instruction Tuning (IT).} In this stage, we maintain the vision encoder in a frozen state, following the approach used in LLaVA models ~\citep{liu2024visual,liu2024improved}. However, unlike the previous stage, we also update the weights of the LLM in addition to tuning the modality projection layer. The objective of this stage is to develop a general-purpose multimodal agent (chatbot) capable of comprehending and executing complex instructions across multiple conversational turns. We describe the datasets used in both the stages in the next section.

\section{Dataset}
\label{sec:dataset}

\begin{figure}
    \centering
    \subfigure[]    {\includegraphics[width=0.4\linewidth]{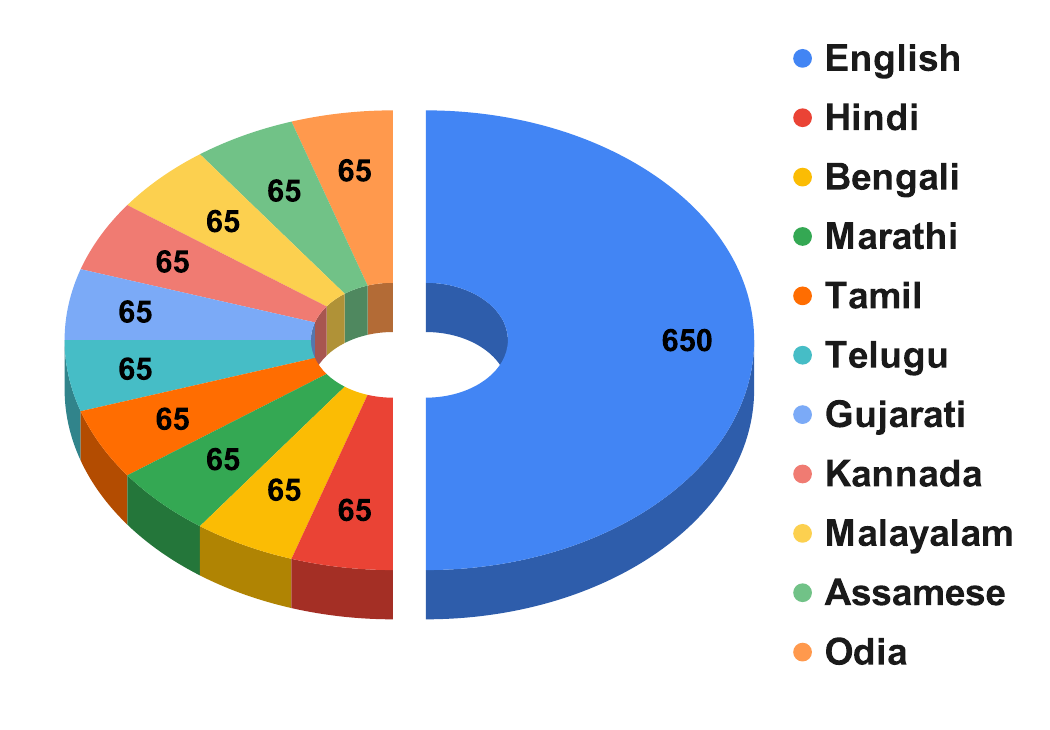}}
    \subfigure[]
    {\includegraphics[width=0.4\linewidth]{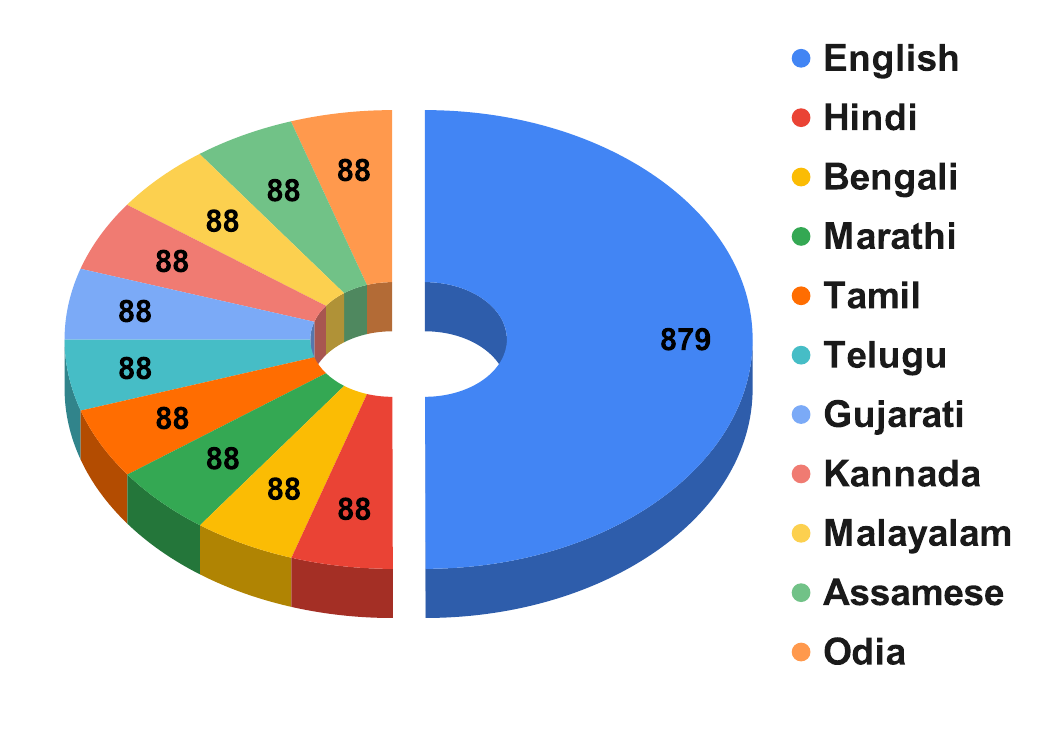}}    
    \caption{\textbf{Language distribution in data mix.} (a) Stage 1 data consists of 1.2M ShareGPT4V in the original English version (650K) and remaining Indian language translations (65K each) (b) Stage 2 data involves 879K samples in English and 88K for each respective language, discussed in Section \ref{sec:dataset}.}
    \label{fig:data}
\end{figure}

Figure \ref{fig:data} illustrates the language distribution of our data mix for both the training stages, which we describe in more detail below: 

\textbf{Stage 1:} For Stage 1 adapter Pre-Training (PT), we use the 1.2 million-sample ShareGPT4V-PT dataset~\citep{chen2023sharegpt4v}, which demonstrated consistent superior performance compared to other PT datasets, such as LLaVA-Pretrain-LCS-558K~\citep{liu2024visual}, in our preliminary experiments. This dataset was subsequently translated into the ten Indic languages supported by the Krutrim LLM. Specifically, we use the open-source model, IndicTrans2~\citep{gala2023indictrans} for this text-only translation task. IndicTrans2 outperformed other translation services (Yandex, ChatGPT, Google Translate, and Bard) in small-scale in-house qualitative human evaluation (win rates 93\% and 80\% for Bengali and Marathi respectively). 
We ensure the pre-training data remained at 1.2M points, with half of the data in English, and sample translations across different languages in an equal ratio to create a balanced multilingual dataset.
This approach was designed to preserve linguistic diversity and computational efficiency, thereby ensuring robust performance in English while developing capabilities in the Indic languages. The balanced dataset mitigates potential biases towards any single language, fostering equitable performance across all supported languages.


\textbf{Stage 2:} The Stage 2 Instruction Tuning (IT) dataset is notably more intricate. The core element of this dataset is the complete English version of LLaVA-1.5-665K~\citep{liu2024improved}. Additionally, we translate LLaVA-Instruct-150K~\citep{liu2024visual} into ten languages using the methodology outlined in Stage 1. Our dataset also incorporates the Cauldron dataset~\citep{laurenccon2024matters}, which includes 50 academic vision-language tasks along with its corresponding in-house translations. Furthermore, we add a substantial collection of images reflecting Indian cultural diversity comprising prominent personalities, monuments, artwork, culinary dishes, and more;  transformed into multilingual pluralistic instruction tuning data, analogous to the open-source English-based LLaVA-IT datasets. Lastly, our dataset features high-quality, text-only English proprietary data. The final composition of the dataset includes approximately 880K English and 90K samples in multiple languages, ensuring a balanced and diverse dataset. This comprehensive range of content supports the development of a model capable of generating and understanding complex descriptions across various domains and visual scenarios, thereby enhancing its reasoning capabilities.


\section{BharatBench Evaluation Suite}
\label{sec:bench}

Although recent efforts have advanced text-only multilingual evaluation~\citep{ahuja2023mega, singh2024indicgenbench}, there is still a lack of evaluation framework for multimodal multilingual scenarios. We introduce BharatBench, a benchmark designed to assess the image understanding capabilities of multilingual Vision-Language Models (VLMs). 
Expanding upon LLaVa-Bench (In-the-Wild) \citep{liu2024visual}, initially adapted for Hindi and Bengali by \citep{maaz2024palo}, we further broadened the benchmark to cover eight additional low resource languages. This extension now forms part of our comprehensive benchmark suite. Furthermore, we include translated versions of prominent VLM evaluation datasets, such as MMVet \citep{yu2023mm} and POPE~\citep{li2023evaluating}
covering all ten languages in our study, in addition to English.




In essence, we intentionally chose to extend existing benchmarks through translation, which 
not only facilitates the creation of valuable multi-way parallel data but also addresses data scarcity issues and leverages the inherent quality of established evaluation frameworks~\citep{singh2024indicgenbench}. This methodology enhances our ability to evaluate and advance multimodal models in a multilingual context. We followed similar guiding principles while creating the training datasets described earlier.

\section{Experiments}
\label{sec:exp}

\subsection{Implementation}

We use PyTorch~\citep{paszke2019pytorch} based HuggingFace Transformers~\citep{wolf2019huggingface} for our experiments. Our Stage 1 and 2 tuning use hyperparameters consistent with those of the LLaVA model~\citep{liu2024visual}, unless otherwise specified. Particularly, we train the model for 1 epoch in both the stages with an overall batch
size of 256 in Stage 1 and 128 in Stage 2. We used cosine LR scheduler with Adam optimizer and a learning rate of 2e-3 and 2e-5 in both the stages respectively. We consider IDEFICS 2 ~\citep{laurenccon2024matters} and PALO~\citep{maaz2024palo} as respective English and multi-lingual baselines and report results from their published work. All our models are trained on 8 $\times$ H100 GPUs which takes around 8 hours for Stage 1 and 18 hours for Stage 2 tuning.  


\begin{figure}[htbp]

\begin{minipage}[b]{1.0\linewidth}
  \centering
\centerline{\includegraphics[scale=0.6]{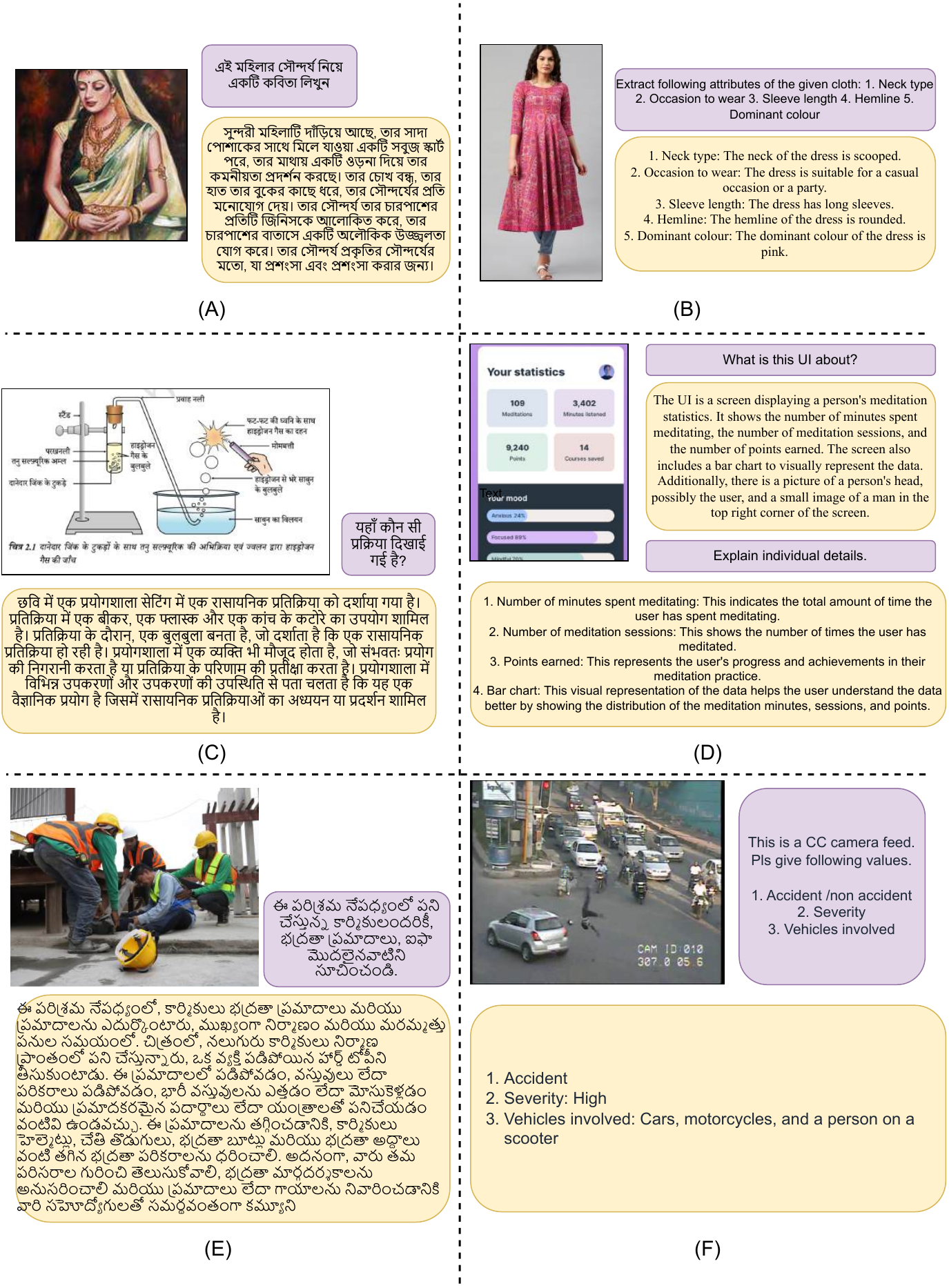}}
\end{minipage}
\caption{\textbf{Multilingual VLM Capabilities.} Our model demonstrates robust performance across various languages in: a) Creative writing, b) Fine-grained attribute extraction, c) Explaining scientific diagrams, d) Screen reading/OCR, e) Anomaly and hazard detection, and f) Real-time accident and incident monitoring.}
    \label{fig:outputs}
\end{figure}


\subsection{English academic benchmarks}

We also evaluate our model using a range of English academic benchmarks, including VQA-v2~\citep{goyal2017making} and GQA~\citep{hudson2019gqa} for visual perception, VizWiz~\citep{gurari2018vizwiz} for zero-shot generalization on questions posed by visually impaired users, and TextVQA~\citep{singh2019towards} for text-rich visual question answering. We also use POPE~\citep{li2023evaluating} to assess hallucination tendencies, MME~\citep{fu2023mme} for yes/no question responses, and LLaVA-Bench (In-the-Wild) \citep{liu2024visual} and MM-Vet \citep{yu2023mm} for visual conversation capabilities. Evaluation scores are reported following prior works.

\begin{figure}[htb]

\begin{minipage}[b]{1.0\linewidth}
  \centering
\centerline{\includegraphics[scale=0.27]{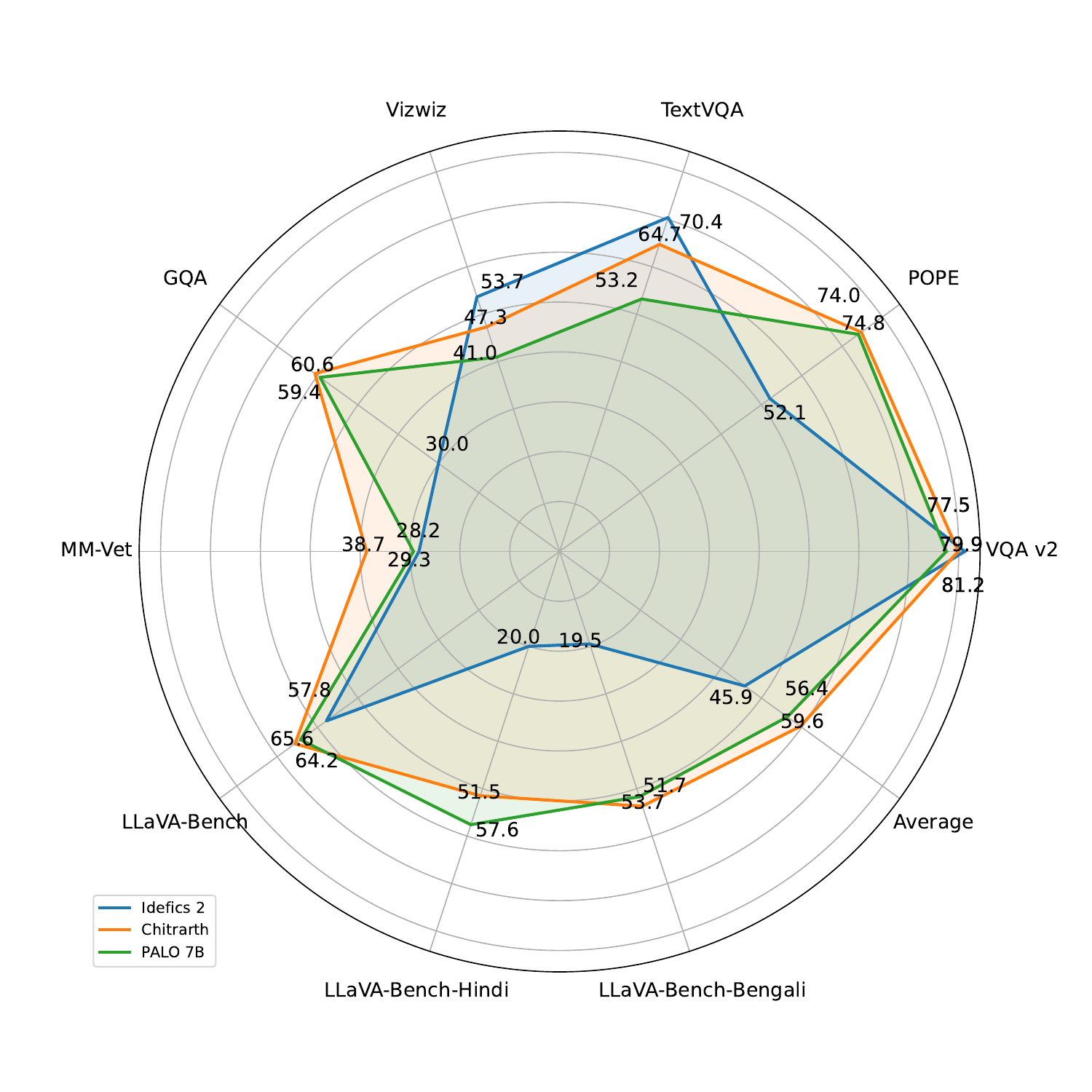}}

\end{minipage}
\caption{\textbf{Performance against SOTA VLMs on different academic multimodal tasks.} Our model consistenly outperforms IDEFICS 2 (7B) and PALO 7B on different benchmarks while remaining competitive on TextVQA and Vizwiz.}
\label{fig:radar_res}
\end{figure}

\begin{table*}[htbp]
\centering
    \resizebox{0.99\linewidth}{!}{
\begin{tabular}{@{}lccccccccccc@{}}
\toprule
\textbf{Bench}       & \multicolumn{1}{l}{\textbf{Telugu}} & \multicolumn{1}{l}{\textbf{Hindi}} & \multicolumn{1}{l}{\textbf{Bengali}} & \multicolumn{1}{l}{\textbf{Malayalam}} & \multicolumn{1}{l}{\textbf{Kannada}} & \multicolumn{1}{l}{\textbf{Assamese}} & \multicolumn{1}{l}{\textbf{Tamil}} & \multicolumn{1}{l}{\textbf{Marathi}} & \multicolumn{1}{l}{\textbf{Gujarati}} & \textbf{Odia} & \multicolumn{1}{l}{\textbf{English}} \\ \midrule
\textbf{POPE}   & 79.9 & 78.68 & 83.24 & 85.29 & 85.52    & 55.59  & 83.28  & 79.17  & 84.75 & 82.03  & 87.63   
\\
\textbf{LLaVA-Bench} & 54.8 & 51.5 & 53.7 & 55.5 & 58.1 & 59.1 & 58.3  & 52.80  & 55.90 & 62.80 & 67.90 \\ 

\textbf{MMVet} & 43.76 & 38.85 & 33.24 & 25.36 & 46.19 & 37.29 & 34.31 & 40.96 & 39.03 & 19.67 & 30.49 \\

\bottomrule
\end{tabular}
}
\caption{\textbf{Performance of Chitrarth on BharatBench Evaluation framework.} Our model is unique in its ability to handle all included languages, setting a baseline for future research.}
\label{tab:indic_evals}

\end{table*}

\begin{figure}[htb]
\begin{minipage}[b]{1.0\linewidth}
  \centering
\centerline{\includegraphics[width=0.6\linewidth]{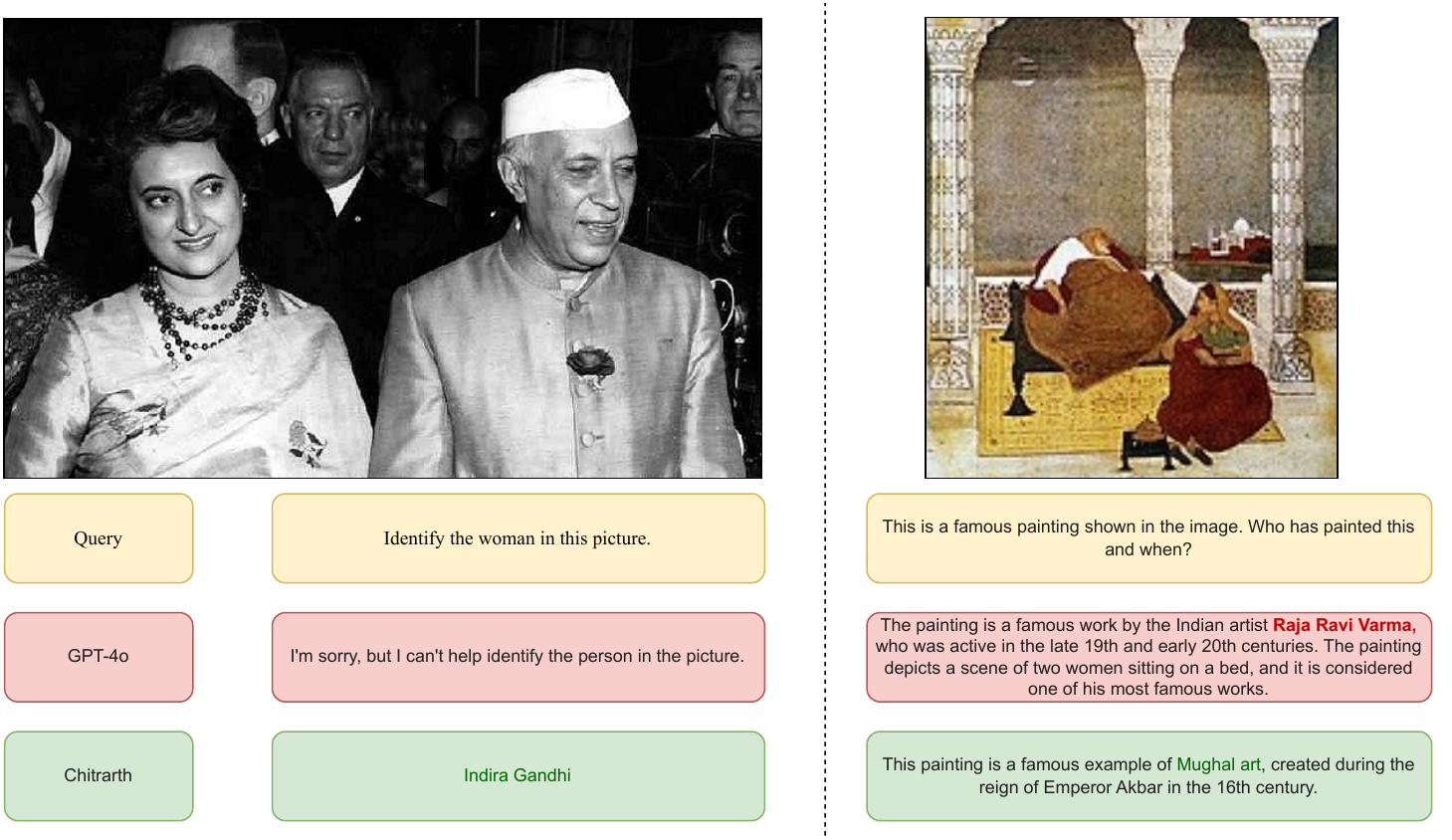}}
\end{minipage}
\caption{\textbf{Performance on images with Indian context.} Chitrarth is able to better understand the images of Image context such as the prominent lady figure (Late Indian Prime Minister Indira Gandhi) in left as well as historical artwork compared to generic and incorrect responses from GPT-4o.}
\label{fig:indic_context}
\end{figure}

\begin{figure}
    \centering
    \subfigure[]{\includegraphics[width=0.49\textwidth]{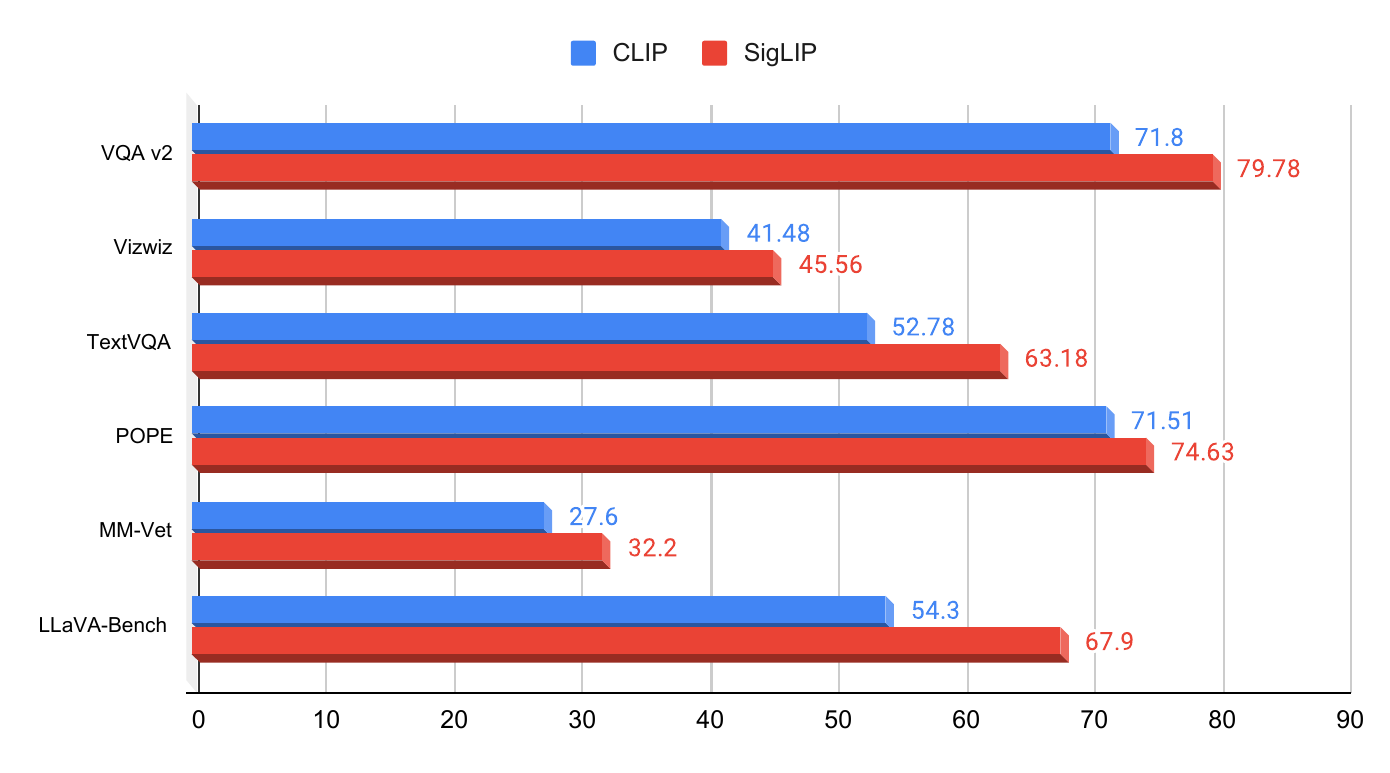}} 
    \subfigure[]
{\includegraphics[scale=0.3]{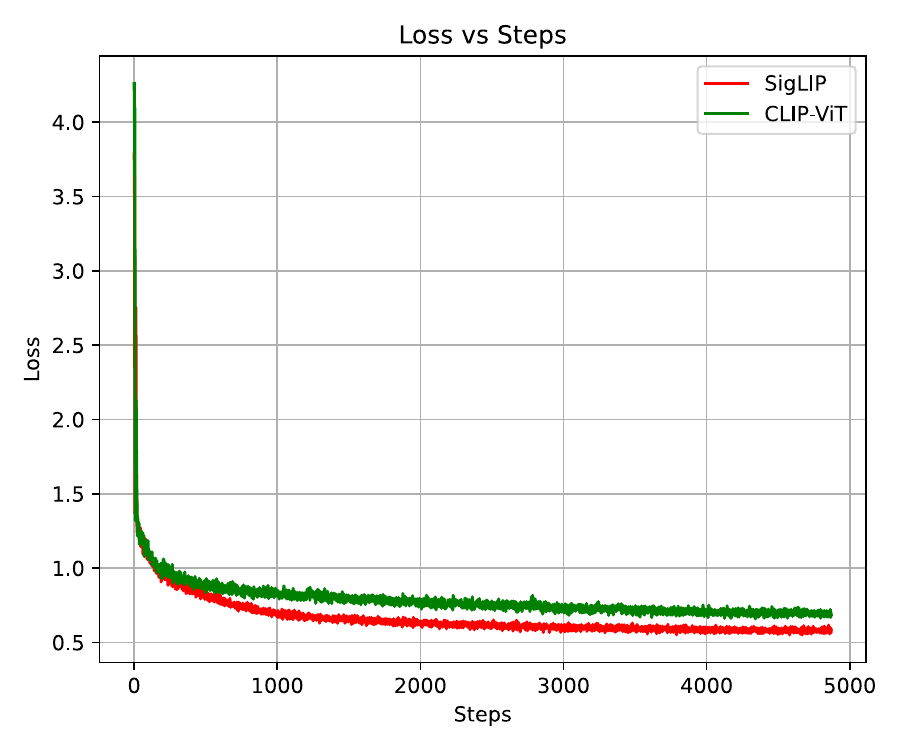}}
    
    \caption{\textbf{Ablation on visual encoder choice.} a) SigLIP as the vision encoder consistently performs better than CLIP in the same training regime. b) SigLIP based model also achieve faster convergence as depicted in Stage 1 loss curve. Stage 2 follows a similar pattern.}
    \label{fig:encoder_ablation}
\end{figure}

\begin{figure}[htb]

  \centering
\centerline{\includegraphics[scale=0.36]{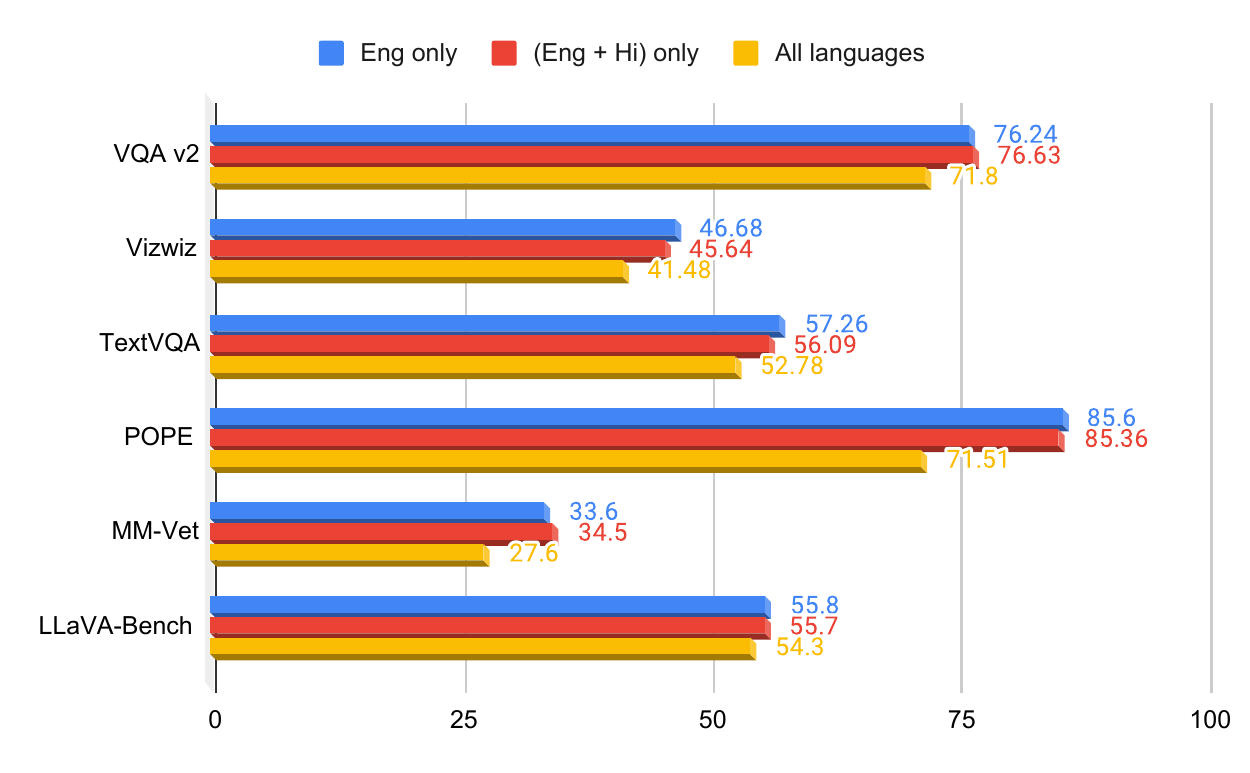}}

%
\caption{\textbf{Impact of Multi-lingual training Data.} Expanding the number of languages in the training data enhances multilingual capabilities but results in decreased scores on academic English datasets.}
\label{fig:data_ablation}

\end{figure}

\subsection{Results}

On the English academic datasets, our model depicts State-of-the-art (SOTA) results for POPE, VQAv2 and GQA compared to the baseline models, while remaining competitive on TextVQA and Vizwiz (see radar graph in Figure \ref{fig:radar_res}). On the LLaVA-Bench (Bengali) our model outperforms the multi-lingual baseline PALO and achieves SOTA results of 53.7 points. Table \ref{tab:indic_evals} presents results on BharatBench across various languages, demonstrating that ours is the only model capable of handling all included languages, establishing baseline results for future research. Figure \ref{fig:outputs} showcases selected outputs from our top-performing Multimodal LLM across various languages. The model excels in tasks such as creative writing, fine-grained attribute extraction, explaining scientific diagrams, and screen reading/OCR, while also demonstrates strong capabilities in anomaly and hazard detection, as well as real-time accident and incident monitoring. In our manual qualitative evaluation, we observe that our model is able to better understand the images of Indian context such as the prominent lady figure in Figure \ref{fig:indic_context}, compared to generic and incorrect responses from GPT-4o. This could be attributed to the inclusion of high quality culturally rich images in Stage 2. A further quantitative analysis around this would be interesting but out of scope of this work.  

We conducted an ablation study evaluating various vision encoders and found that SigLIP-SO400M consistently outperforms CLIP ViT-L/14@336px across all English benchmarks, achieving faster convergence (see Figure \ref{fig:encoder_ablation}
). Notably, SigLIP-SO400M yields improvements of 11 points on TextVQA and 13 points on LLaVA-Bench compared to CLIP ViT-L/14@336px. Figure \ref{fig:data_ablation} explores the impact of multilingual training data on the English academic benchmarks. We compare our model's performance when trained with only English, bilingual, and multilingual data across both stages. Consistent with the findings of ~\citep{scao2022language}, expanding the range of languages in the training data improves multilingual capabilities but leads to decreased performance on academic English datasets. This underscores a key challenge in balancing cross-lingual performance.

\section{Conclusion}
\label{sec:conclusion}
This paper presents Chitrarth, a multilingual multimodal LLM that is able to have image grounded conversations in English as well as across multiple Indian languages. Our model encodes images using a pre-trained vision encoder~\citep{alexey2020image} and autoregressively generates response using a pre-trained multi-lingual LLM. Empirically, our model outperforms previous baselines for different multimodal tasks. As part of this work, we also introduce BharatBench, a multimodal evaluation framework and provide benchmark results for low resource languages. We anticipate that our research will significantly contribute to the advancement of VLMs for Indian languages, thereby providing substantial benefits to a population exceeding one billion people. 

\textbf{Limitations and Future Work:}
We use an automated translation pipeline for creating multi-lingual training data
which may introduce biases from large language models (LLMs), potentially leading to misrepresentations of cultural symbols and gestures, impacting content accuracy. Addressing these biases requires additional evaluation and targeted training, which we plan to address in the future work. Building on our promising results across 10 low-resource languages, we plan to broaden the language scope in the future research to enhance linguistic diversity and inclusivity in our Vision-Language Models (VLMs). In our current training pipeline, we keep the vision encoder frozen throughout both training stages. However, recent research \citep{laurenccon2024matters, tong2024cambrian} suggests that unfreezing the vision encoder could enhance representation learning. We plan to investigate this approach in future work with higher resolution vision encoders, along with expanding our model's ability to interpret multiple images within a conversational context.
 

\section*{Acknowledgements}
We thank Bhavish Aggarwal and the rest of the Krutrim team which helped with model development at various stages. Our models were trained with generous support from Krutrim cloud using Krutrim credits. We also thank the reviewers for their valuable feedback and suggestions.

\bibliographystyle{plainnat}
\bibliography{anthology-neurips}












\newpage

\end{document}